\documentclass{article}

\usepackage{spconf}

\usepackage[utf8]{inputenc} 
\usepackage[T1]{fontenc}    
\usepackage{url}            
\usepackage{booktabs}       
\usepackage{amsfonts}       
\usepackage{nicefrac}       
\usepackage{microtype}      %
\usepackage{enumitem,ragged2e}
\usepackage{algorithm,algpseudocode}
\usepackage{amsmath,mathtools,amsthm,graphicx,tabularx,epstopdf,url}

\makeatletter
\g@addto@macro\normalsize{%
  \setlength\abovedisplayskip{2pt}
  \setlength\belowdisplayskip{2pt}
  \setlength\abovedisplayshortskip{1pt}
  \setlength\belowdisplayshortskip{1pt}
}
\makeatother

\usepackage{xcolor}

\def\bb{\mathbf{b}}

\def\bx{\mathbf{x}}

\def\bPhi{\boldsymbol{\Phi}}

\def\bmu{\boldsymbol{\mu}}

\def\rone{\overrightarrow{\mathbf{1}}}

\def\bI{\mathbf{I}}
\def\bA{\mathbf{A}}
\def\bB{\mathbf{B}}
\def\bC{\mathbf{C}}
\def\bG{\mathbf{G}}
\def\bM{\mathbf{M}}
\def\bU{\mathbf{U}}
\def\bW{\mathbf{W}}
\def\bX{\mathbf{X}}
\def\bY{\mathbf{Y}}

\def\b0{\mathbf{0}}
\def\Sw{{\mathbf{S}_W}}
\def\Sb{{\mathbf{S}_B}}
\def\bK{\mathbf{K}}
\def\Kbar{\bar{\mathbf{K}}}
\def\Kb{{\mathbf{K}_B}}
\def\Sbar{\bar{\mathbf{S}}}

\def\bSigma{\boldsymbol{\Sigma}}

\def\lnorm{\left\|}
\def\rnorm{\right\|}
\def\lp{\left(}
\def\rp{\right)}

\DeclareMathOperator{\trace}{tr}

\theoremstyle{plain}

\usepackage{color}

\newcolumntype{Y}{>{\centering\arraybackslash}X}
\allowdisplaybreaks

\title{Scalable Kernel Learning via the Discriminant Information}

\name{Mert Al,\qquad Zejiang Hou,\qquad Sun-Yuan Kung \sthanks{Copyright 2020 IEEE. Published in the IEEE 2020 International Conference on Acoustics, Speech, and Signal Processing (ICASSP 2020), scheduled for 4-9 May, 2020, in Barcelona, Spain. Personal use of this material is permitted. However, permission to reprint/republish this material for advertising or promotional purposes or for creating new collective works for resale or redistribution to servers or lists, or to reuse any copyrighted component of this work in other works, must be obtained from the IEEE. Contact: Manager, Copyrights and Permissions / IEEE Service Center / 445 Hoes Lane / P.O. Box 1331 / Piscataway, NJ 08855-1331, USA. Telephone: + Intl. 908-562-3966.}}
\address{Princeton University}

\begin{document}
\ninept

\maketitle

\begin{abstract}
Kernel approximation methods create explicit, low-dimensional kernel feature maps to deal with the high computational and memory complexity of standard techniques. This work studies a supervised kernel learning methodology to optimize such mappings. We utilize the Discriminant Information criterion, a measure of class separability with a strong connection to Discriminant Analysis. By generalizing this measure to cover a wider range of kernel maps and learning settings, we develop scalable methods to learn kernel features with high discriminant power. Experimental results on several datasets showcase that our techniques can improve optimization and generalization performances over state of the art kernel learning methods.
\end{abstract}

\begin{keywords}
Kernel learning, discriminant analysis, scalable learning, kernel approximation, classification
\end{keywords}

\section{Introduction}
\vspace{-2pt}
The main innovation of kernel methods is the mapping of the data onto a high-dimensional feature space without having to compute the expansions explicitly \cite{vapnik2013nature,kung2014kernel}. This is achieved via the kernel trick, which only requires a Gram (kernel) matrix to be computed in the original feature space. Given $N$ training samples, the kernel matrix is $N \times N$. Thus, although this may be advantageous for small-scaled applications,
for large-scaled learning \textendash{} where $N$ can be massive \textendash{} the size of the kernel matrix quickly becomes an obstacle. One of the prominent approaches to address this challenge is kernel matrix approximation \cite{girolami2002orthogonal,drineas2005nystrom,rahimi2008random}. These methods lead to explicit, low-dimensional, approximate representations of the implicit, high-dimensional mappings for the data, which can then be used in large-scaled learning applications. 

Kernel approximation methods are most commonly a variant of the data dependent Nystr\"om method \cite{girolami2002orthogonal,drineas2005nystrom} or the data independent Random Fourier method \cite{rahimi2008random,rahimi2009weighted}. A recent approach to kernel learning involves the parametrization and subsequent optimization of these kernel mappings \cite{yang2015carte,si2016goal}, and previous work has achieved impressive results through gradient based training of kernel machines \cite{dai2014scalable,lu2014scale}. In a similar vein, our work optimizes over parametric families of kernels using gradient methods, but distinctively, we optimize based on explicit measures of the discriminant power of kernel maps. 

Due to the relation of this objective to \emph{Discriminant Analysis}, we propose the \emph{Discriminant Information} (DI) criterion \cite{kung2017compressive} to optimize the Random Fourier features, and we derive an equivalent criterion for the Nystr\"om features. The proposed kernel optimization method can search for a kernel feature subspace with high discriminant power without suffering the high computational and memory cost of traditional Kernel Discriminant Analysis. Additionally, it is suitable for both classification and regression settings. Our experimental results on several datasets demonstrate that, not only can the proposed training methodology lead to better objective values compared to existing algorithms, but it can also lead to models that generalize better to unseen examples despite similar fits on the training set.

\vspace{-5pt}
\section{Related Work}
\vspace{-2pt}
The standard Nystr\"{o}m algorithm can be viewed as the application of KPCA to a small, randomly chosen subset of training samples \cite{girolami2002orthogonal}. Various works have altered this method to achieve better approximations with less memory/computation. Zhang \textit{et al.} use k-means centroids to perform KPCA instead of a random subset of the data \cite{zhang2008improved}. Kumar \textit{et al.} combine multiple smaller scale KPCAs \cite{kumar2009ensemble}. Li \textit{et al.} utilize randomized SVD to speed up KPCA for the Nystr\"{o}m algorithms \cite{li2015large}. Additionally, non-uniform sampling schemes have been explored to improve the memory performance of Nystr\"{o}m \cite{gittens2016revisiting}.

Many data independent approximations of kernel based features have also been proposed. Rahimi and Recht introduced random features to approximate shift invariant kernels \cite{rahimi2008random,rahimi2009weighted}. The most well-known of such techniques is the \emph{Random Fourier Features}. 
These methods were later extended to achieve improved complexity and versatility. For example, Le \textit{et al.} approximate the frequency sampling step via a series of cheap matrix products \cite{le2013fastfood}, and Yang \textit{et. al} further optimize the resulting feature mappings via Gaussian Processes \cite{yang2015carte}. 

A more recent data dependent approach 
involves optimizing these low-dimensional kernel maps based on an objective function. In \cite{mairal2014convolutional}, a squared error criterion is used to approximate a kernel mapping within multiple layers of a Convolutional Neural Network (CNN). An alternating minimization approach is proposed in \cite{si2016goal}, where the kernel mapping is trained jointly with a linear model for Inductive Matrix Completion, and extensions of this methodology are also presented for multiclass/multilabel learning and semi-supervised clustering. 

We follow the optimization approach in this paper based on supervised learning criteria, however, in lieu of training linear models, we directly apply measures of discriminant power to Nystr\"om and Random Fourier features. Furthermore, we only apply these measures with small subsets (i.e. mini-batches) of the data during training. 
\vspace{-5pt}
\section{Preliminaries}
\vspace{-2pt}

\setlength{\belowdisplayskip}{0pt} \setlength{\belowdisplayshortskip}{0pt}
\setlength{\abovedisplayskip}{0pt} \setlength{\abovedisplayshortskip}{0pt}

 \textit{Notation:} We denote by $\bX$ an $N$-columned data matrix and by $\bY$ an $N$-rowed target matrix. $\bK$ denotes the $N \times N$ kernel matrix and $\bPhi$ denotes the $N$-columned data matrix in the kernel induced feature space.
 $\bC=\bI-\frac{1}{N}\mathbf{1}$ denotes the data centering matrix. 
 For a matrix $\bM$, we denote its best rank-$k$ approximation by $\bM_{[k]}$, its Moore-Penrose inverse by $\bM^{+}$ and its Frobenius norm by $\lnorm \bM \rnorm_F$. With slight abuse of notation, we denote by $k(\bX_1,\bX_2)$ the $N_1 \times N_2$ kernel matrix that results from evaluating the kernel function $k(\cdot,\cdot)$ on the $N_1$ and $N_2$-columned data matrices $\bX_1$ and $\bX_2$. 
\vspace{-5pt}
\subsection{Discriminant Analysis}
\vspace{-2pt}
The Linear Discriminant Analysis (LDA) projects $L$ classes of data onto $L-1$ directions in such a way that maximizes the ratio of between class separation and within class separation \cite{bryan1951generalized}. These directions can be found by solving the optimization problem
\begin{equation}
 \underset{\bW}{\text{maximize}} \frac{\left| \bW^\top \Sb \bW \right|}{\left|\bW^\top \Sw \bW \right|} \text{,}
\label{eq:LDA}
\end{equation}
where $\Sw = \sum_{c=1}^L \sum_{x_i \in c} \lp \bx_i-\bmu_c \rp \lp \bx_i-\bmu_c \rp^T$ and $\Sb = \sum_{c=1}^L N_c \lp \bmu-\bmu_c \rp \lp \bmu-\bmu_c \rp^T$ are the within and between-class scatter matrices, respectively, with $\bmu$ being the dataset mean, $\bmu_c$ the class mean, and $N_c$ the number of samples in class $c$. LDA was later extended to Kernel Discriminant Analysis (KDA), which projects the data onto the $L-1$ directions in a kernel induced feature space \cite{roth2000nonlinear}.

Kung more recently proposed a related optimization objective called \emph{Discriminant Component Analysis} (DCA) \cite{kung2017discriminant} defined as
\begin{equation}
    \underset{\bW \colon \bW^\top \lp \Sbar+\rho \bI \rp \bW = \bI}{\text{maximize}} \trace \lp \bW ^\top \Sb \bW \rp\text{,}
    \label{eq:DCA}
\end{equation}
where $\Sbar=\Sb+\Sw$ is the scatter matrix, and $\rho\bI$ is a regularizer. 

When $\rho=0$, the optimal solution of DCA is also optimal for LDA.
Moreover, when $\rho>0$ and every data sample is considered as the sole member of its own class ($\Sb=\Sbar$), the optimal DCA solution is also the optimal Principal Component Analysis (PCA) solution up to the scaling of the columns of $\bW$. Hence, DCA can be considered as a supervised generalization of PCA. The kernelized version of this objective called Kernel DCA (KDCA) was also presented in \cite{kung2017discriminant}, 
\begin{equation}
    \underset{\bA \colon \bA^\top \lp \Kbar^2+\rho \Kbar \rp \bA = \bI}{\text{maximize}} \trace \lp \bA ^\top \Kb \bA \rp\text{,}
    \label{eq:KDCA}
\end{equation}
where $\Kbar=\bC\bK\bC$ is the centered kernel matrix and $\Kb$ is a kernelized counterpart of $\Sb$. KDCA can in turn be considered as a supervised generalization of Kernel PCA (KPCA).

Methods like KPCA and KDCA are suitable for extracting low-dimensional kernel mappings. However, they suffer from $O(N^3)$ computational complexity. Hence, it is necessary to combine these objectives with kernel approximation methods to scale to large datasets. 
\vspace{-5pt}
\subsection{Kernel Approximation}
\label{sec:ker_app}
\vspace{-2pt}
Although our method can be extended to various types of kernel maps, we consider two particular methods, which are summarized below.

\emph{Nystr\"{o}m} \cite{girolami2002orthogonal} method projects the data into a kernel induced feature subspace spanned by $n \ll N$ representative data points. Using $\bX_r$ and $\bPhi_r$ to denote the representative data points and their feature space projections, respectively, the resulting rank-$k$ approximation of the kernel matrix is given by $\widetilde{\bK}=\bG\bB_{[k]}^+\bG^\top$, where $\bG=\bPhi^\top \bPhi_r=k(\bX,\bX_r)$ and $\bB=\bPhi_r^\top \bPhi_r=k(\bX_r,\bX_r)$. This is equivalent to applying the feature mapping to the training data; $\phi(\bX;\bX_r)=\bSigma^{-\nicefrac{1}{2}}\bU^\top k^\top(\bX,\bX_r)$, where $\bU\bSigma\bU^\top$ is the compact SVD of $\bB_{[k]}$. Usually, the representative data points are obtained via random sampling or k-means clustering of the training samples.
 
 \emph{Random Fourier} \cite{rahimi2008random} method approximates a kernel mapping by sampling components from the Fourier transform of a (shift-invariant) kernel function. An approximation of this variant can be obtained via the transformation $\phi(\bX;\bW_f,\bb_f)=\sqrt{\nicefrac{2}{J}}\cos({\bW_f^\top\bX+\bb_f\rone^\top})$, where $\bW_f$ is sampled from the Fourier transform of the kernel function and $\bb_f$ is sampled uniformly from $[ 0, 2\pi ]$. 

\vspace{-5pt}
\section{Methodology}
\vspace{-2pt}
\subsection{The Discriminant Information Criterion} 
\vspace{-2pt}
\label{subsec:DI}
The Discriminant Information (DI) criterion \cite{kung2017compressive} is closely related to LDA, DCA and Ridge Regression (RR) \cite{friedman2001elements}. To establish this connection, let us first write the Kernel RR (KRR) objective,
\begin{equation}
 \underset{\bW,\bb}{\text{minimize}} \left\| \bPhi^\top\bW+\rone\bb^\top-\bY \right\|_F^2+\rho\left\|\bW\right\|_F^2 \text{.}
\label{eq:RR}
\end{equation}
Setting the gradients equal to zero yields the optimal bias vector $\bb^*=N^{-1}\lp \bY^\top \rone-\bW^\top \bPhi \rone \rp$ and the optimal weight matrix $\bW^* = \lp \Sbar+\rho\bI \rp^{-1}\bar{\bPhi}\bar{\bY}$, with $\Sbar=\bar{\bPhi}\bar{\bPhi}^\top$, $\bar{\bPhi}=\bPhi \bC$, $\bar{\bY} = \bC \bY$. Notice that $\bar{\bPhi}\bar{\bY}=\bPhi \bC \bC \bY=\bPhi \bC \bY=\bar{\bPhi}\bY$. Upon plugging in the optimal solution to the objective in \eqref{eq:RR}, we can express the minimum regularized least squares error (MRLSE) as
\begin{equation}
\text{MRLSE} = -\trace\lp \lp \Sbar+\rho \bI \rp^{-1} \Sb \rp + \lnorm\bar{\bY}\rnorm_F^2 \text{,}
\end{equation}
where $\Sb=\bar{\bPhi}\bY \bY^{\top}\bar{\bPhi}^\top$. $\Sb$ is the same as the previously defined between-class scatter matrix when $\bY$ is a class indicator matrix with each column scaled to be unit norm. However, this new definition naturally encompasses the regression setting with arbitrary $\bY$, therefore, we will use it for the rest of this paper. Ignoring the constant term, we see that MRLSE can be minimized by maximizing the quantity we refer to as \emph{the Discriminant Information} (DI),
\begin{equation}
    \text{DI} = \trace\lp \lp \Sbar+\rho \bI \rp^{-1} \Sb \rp\text{.}
    \label{eq:DI}
\end{equation}
This is a natural multiclass/multilabel extension of \emph{Fisher Discriminant Ratio} (FDR), a useful measure of class separability \cite{kung2014kernel,mao2002rbf,eric2008testing}.

We can plug in the Random Fourier (RF) feature mapping $\phi(\bX; \bW_f,\bb_f)=\sqrt{\nicefrac{2}{J}}\cos({\bW_f^\top\bX+\bb_f\rone^\top})$ directly into \eqref{eq:DI}, resulting in the objective function
\begin{multline}
    \text{RFDI} \lp \bX,\bY;\boldsymbol{\theta} \rp = \trace \biggl( \lp \phi(\bX;\boldsymbol{\theta})\bC \phi^\top(\bX;\boldsymbol{\theta})+\rho \bI \rp^{-1}  \\  \phi(\bX;\boldsymbol{\theta}) \bC\bY \bY^\top \bC \phi^\top(\bX;\boldsymbol{\theta}) \biggr) \text{,}
\label{eq:RF_obj}
\end{multline}
with $\boldsymbol{\theta} \coloneqq (\bW_f,\bb_f)$ representing the optimization parameters.


To use DI as an optimization objective with Nystr\"om features, we incorporate the orthogonalization procedure into this metric. We start by expressing DI as the maximal objective value of DCA.
\begin{align}
\text{DI} &= \trace\lp \lp \Sbar+\rho \bI \rp^{-\frac{1}{2}} \Sb \lp \Sbar+\rho \bI \rp^{-\frac{1}{2}} \rp \nonumber \\
 &= \underset{\widehat{\bW} \colon \widehat{\bW}^\top \widehat{\bW} = \bI}{\text{max}} \trace\lp \widehat{\bW}^\top \lp \Sbar+\rho \bI \rp^{-\frac{1}{2}} \Sb \lp \Sbar+\rho \bI \rp^{-\frac{1}{2}} \widehat{\bW} \rp \nonumber \\
 &= \underset{\bW \colon \bW^\top \lp \Sbar+\rho \bI \rp \bW = \bI}{\text{max}} \trace \lp \bW ^\top \Sb \bW \rp \text{,}
\label{eq:alt_DI}
\end{align}
where the second equality is due to the trace of a symmetric matrix being equal to the sum of its eigenvalues, and the third equality is due to change of variables $\bW=\lp \Sbar+\rho \bI \rp^{-\frac{1}{2}}\widehat{\bW}$.

Since the Nystr\"om projection places the data $\bPhi$ in the span of $\bPhi_r$, there exists a maximizer of the DCA objective in \eqref{eq:alt_DI} that satisfies $\bW = \bPhi_r \bA$ for some matrix $\bA$. By using this observation with the kernel trick we obtain $\bW^\top \Sbar \bW = \bA^\top \bG^\top \bC \bC \bG \bA$, $\bW^\top \Sb \bW = \bA^\top \bG^\top \bC \bY \bY^\top \bC \bG \bA$, and $\bW^\top\bW = \bA^\top \bB \bA$. By plugging these results into \eqref{eq:alt_DI}, we derive an equivalent expression for the Nystr\"om (Nys) features, which we name \emph{the Kernel DI} (KDI),
\begin{equation}
\begin{aligned}
    \text{KDI} &= \underset{\bA \colon \bA^\top \lp \bar{\bG}^\top \bar{\bG}+\rho \bB \rp \bA = \bI}{\text{max}} \trace \lp \bA ^\top \bar{\bG}^\top \bY \bY^\top \bar{\bG} \bA \rp \\
    &= \trace \lp \lp \bar{\bG}^\top \bar{\bG}+\rho \bB \rp^{+} \bar{\bG}^\top \bY \bY^\top \bar{\bG} \rp \text{,}
\end{aligned}
\label{eq:KDI}
\end{equation}
where $\bar{\bG}=\bC \bG$ and the proof of the second equality is analogous to \eqref{eq:alt_DI}. Further plugging in $\bG=k(\bX,\bX_r)$ and $\bB=k(\bX_r,\bX_r)$ into \eqref{eq:KDI} results in the objective function
\begin{multline}
    \text{NysDI}(\bX,\bY;\boldsymbol{\theta})=\trace \Bigl( \lp k^\top(\bX,\boldsymbol{\theta})\bC k(\bX,\boldsymbol{\theta})+\rho k(\boldsymbol{\theta},\boldsymbol{\theta}) \rp^{+} \\ k^\top(\bX,\boldsymbol{\theta}) \bC \bY \bY^\top \bC k(\bX,\boldsymbol{\theta}) \Bigr)\text{,}
\label{eq:Nys_obj}
\end{multline}
with $\boldsymbol{\theta} \coloneqq \bX_r$ representing the optimization parameters.


If we allow the representative sample size to be as high as the number of training samples ($n=N$), we can optimally set $\bX_r=\bX$, yielding $\bG=\bB=\bK$. This reduces KDI to KDCA, however, it makes KDI computation an $O(N^3)$ operation. When $n \ll N$, the alternative KDCA solution in \eqref{eq:KDI} can be computed in $O(N n^2)$, but the solution is restricted to a small subspace. Consequently, KDI based kernel optimization can be thought of as a way to search for the best $n$-dimensional kernel subspace to efficiently perform KDCA.

\vspace{-5pt}
\subsection{The Optimization Procedure}
\vspace{-2pt}
\begin{algorithm}[t]
\caption{KDI Based Nystr\"om Feature Optimization}
\begin{algorithmic}
\footnotesize
\State \textbf{Input:} Training data: $(\bX,\bY)$; model parameters: $\rho$, $J$, the kernel parameters; and $batch\_size$.
\State Initialize $J$ representative samples $\bX_r$ by randomly sampling or k-means clustering the training samples $\bX$
 \Repeat
  \For {$b=1, \ldots, \lfloor N/batch\_size \rfloor$} 
   \State Extract a mini-batch $(\bX',\bY') \subset (\bX,\bY)$\;
   \State Update $\boldsymbol{\theta}=\bX_r$ via its gradient w.r.t. $\text{NysDI}(\bX',\bY';\boldsymbol{\theta})$ \eqref{eq:Nys_obj}
   \State $\mu' \gets \frac{b-1}{b}\mu'+ \frac{1}{b}\text{NysDI}(\bX',\bY';\boldsymbol{\theta})$
  \EndFor
  \State $\mu_\text{KDI} \gets \mu'$
 \Until $\mu_\text{KDI}$ converges
\State \textbf{Output:} Optimized representative samples $\bX_r$.
\end{algorithmic}
\label{alg:Nystrom}
\vspace{-3pt}
\end{algorithm}

\begin{algorithm}[t]
\caption{DI Based Fourier Feature Optimization}
\begin{algorithmic}
\footnotesize
\State \textbf{Input:} Training data: $(\bX,\bY)$; model parameters: $\rho$, $J$, the kernel parameters; and $batch\_size$.
\State Initialize $(\bW_f, \bb_f)$ by sampling $\bW_f$ from the Fourier transform of the kernel function and $\bb_f$ uniformly from $[ 0, 2\pi ]$
 \Repeat
  \For {$b=1, \ldots, \lfloor N/batch\_size \rfloor$} 
   \State Extract a mini-batch $(\bX',\bY') \subset (\bX,\bY)$\;
   \State Update $\boldsymbol{\theta} = [\bW_f^\top\ \bb_f]$ via its gradient w.r.t. $\text{RFDI}(\bX',\bY';\boldsymbol{\theta})$ \eqref{eq:RF_obj}
   \State $\mu' \gets \frac{b-1}{b}\mu'+ \frac{1}{b}\text{RFDI}(\bX',\bY';\boldsymbol{\theta})$
  \EndFor
  \State $\mu_\text{DI} \gets \mu'$
 \Until $\mu_\text{DI}$ converges
\State \textbf{Output:} Optimized Fourier feature parameters $(\bW_f, \bb_f)$.
\end{algorithmic}
\label{alg:Fourier}
\vspace{-3pt}
\end{algorithm}

\label{subsec:optimization}
Both the NysDI in \eqref{eq:Nys_obj} and RFDI in \eqref{eq:RF_obj} are differentiable with respect to the optimization parameters $\boldsymbol{\theta}$ as long as $k(\cdot,\boldsymbol{\theta})$ and $\phi(\cdot;\boldsymbol{\theta})$ are differentiable with respect to $\boldsymbol{\theta}$. This requirement is satisfied by a wide range of kernel functions $k$ and feature maps $\phi$, though, we restrict ourselves to Nys and RF features in this paper. Our gradient based optimization method is summarized in Algorithms \ref{alg:Nystrom} and \ref{alg:Fourier}.  

We parametrize Nys features by the representative data points $\bX_r$ and RF features by the linear projection weights and bias ($\bW_f,\bb_f$). We apply NysDI and RFDI gradients to optimize Nys and RF features, respectively (via gradient ascent), and we stop the training when the average mini-batch NysDI/RFDI saturates. As a regularizer, we keep the batch sizes larger than the feature dimensionalities.

With batch size $N_b$ and feature dimensionality $J$, the proposed optimization strategies require only $O(N_b J+J^2)$ memory and lead to $O(N_b J^2 + J^3)$ computational cost per iteration. For commonly used kernels such as Gaussian, the gradient computations mainly consist of matrix products and linear system solutions, which can be sped up significantly with GPU-accelerated linear system solvers. For instance, our implementation took less than 80 miliseconds to compute NysDI/RFDI gradients on an nVidia P100 GPU with $N_b \leq 4000$ and $J \leq 2000$ using Gaussian kernels on the presented datasets.





\section{Experiments}
\vspace{-2pt}
\subsection{Experimental Setup}
\vspace{-5pt}

\begin{table}[t]
  \caption{Summary of the datasets used in the experiments.}
  \vspace{2mm}
  \footnotesize
  \centering
  \begin{tabularx}{\linewidth}{l Y Y Y Y Y}
  \toprule
  Dataset & \# Feat. & \# Train & \# Test & \# Class \\
  \midrule
   Letter \cite{frey1991letter} & $16$ & $15000$ & $5000$ & $26$ \\ 
   MNIST \cite{lecun1998gradient} & $784$ & $60000$ & $10000$ & $10$ \\
   CovType \cite{blackard1999comparative} & $54$ & $464810$ & $116202$ & $7$ \\
   \bottomrule
  \end{tabularx}
  \label{tab:Datasets}
  \vspace{-5pt}
\end{table}

\begin{figure}[t]
\hspace{1.3cm}
\begin{minipage}[t]{0.1\textwidth}
  \centering
  \centerline{\includegraphics[width=4.5cm]{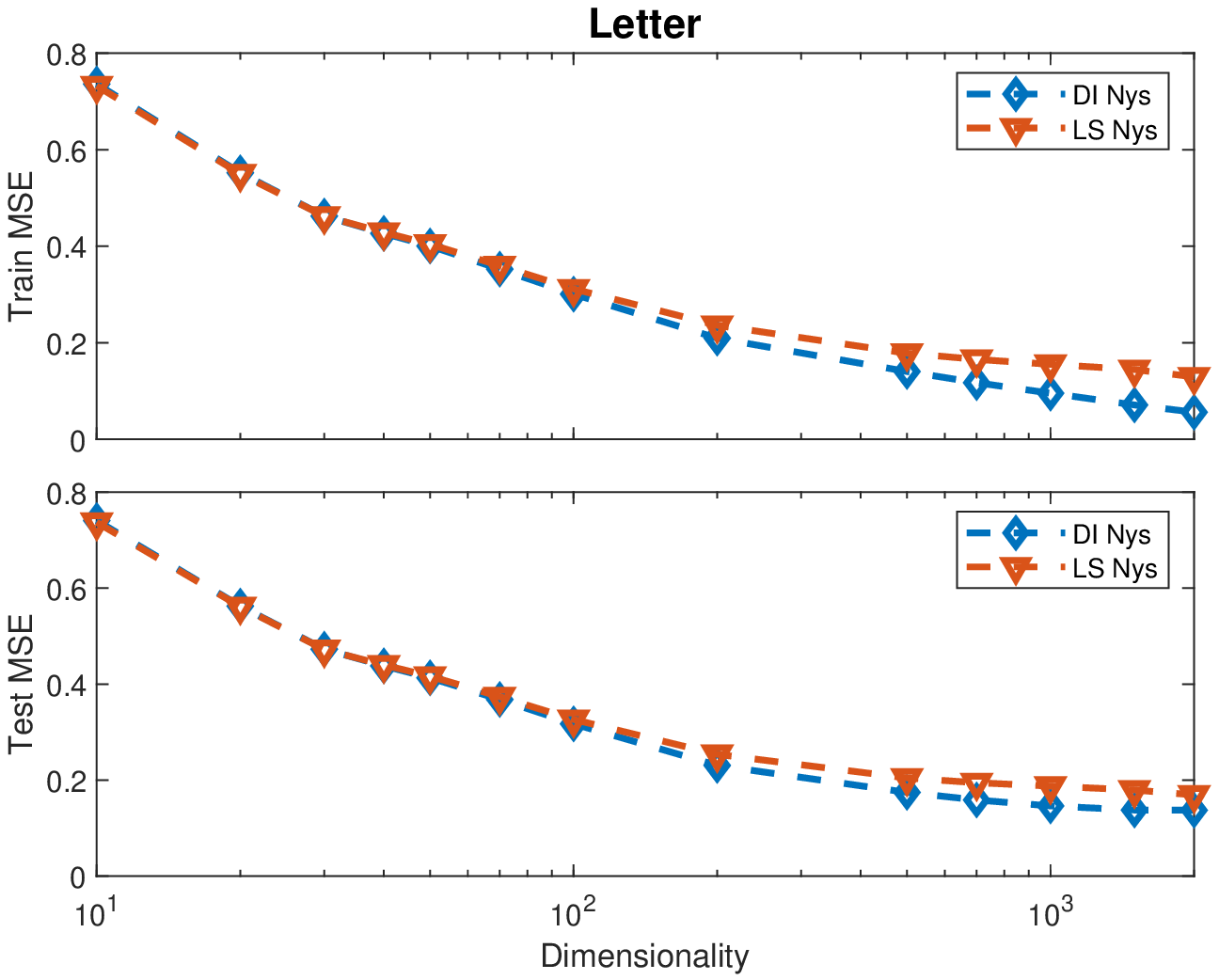}}
\end{minipage}
\hspace{2.5cm}
\begin{minipage}[t]{0.1\textwidth}
  \centering
  \centerline{\includegraphics[width=4.5cm]{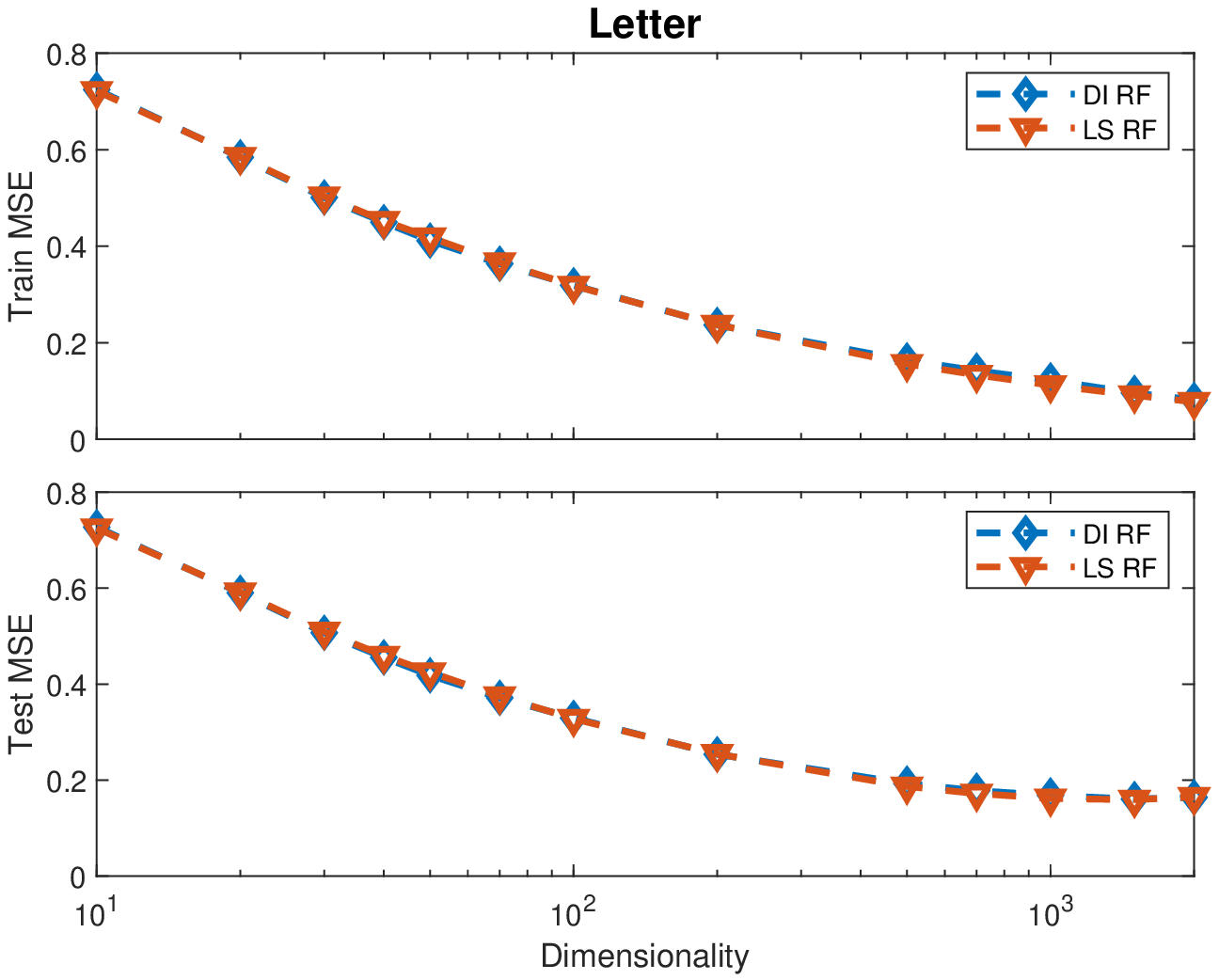}}
\end{minipage}

\hspace{1.3cm}
\begin{minipage}[t]{0.1\textwidth}
  \centering
  \centerline{\includegraphics[width=4.5cm]{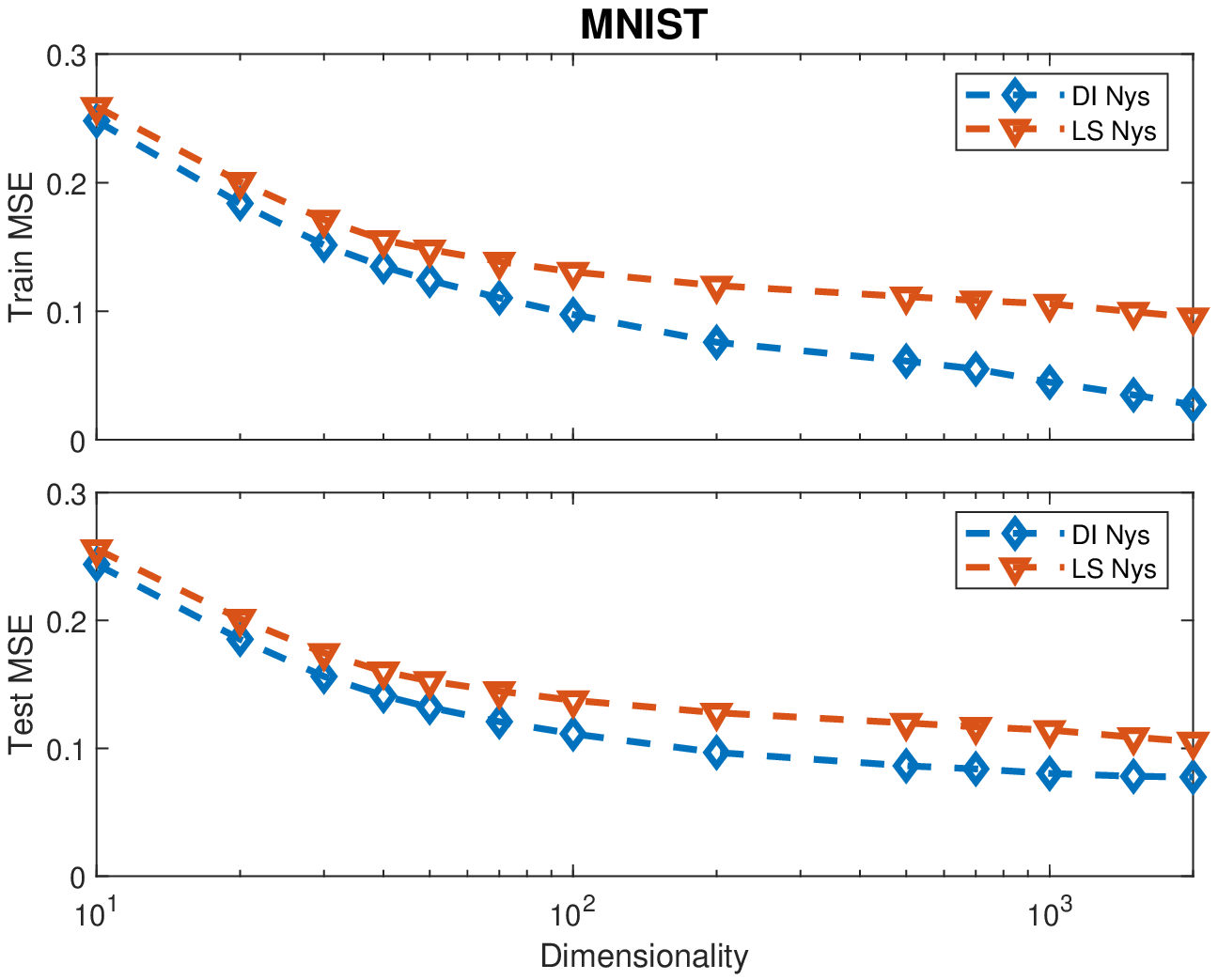}}
\end{minipage}
\hspace{2.5cm}
\begin{minipage}[t]{0.1\textwidth}
  \centering
  \centerline{\includegraphics[width=4.5cm]{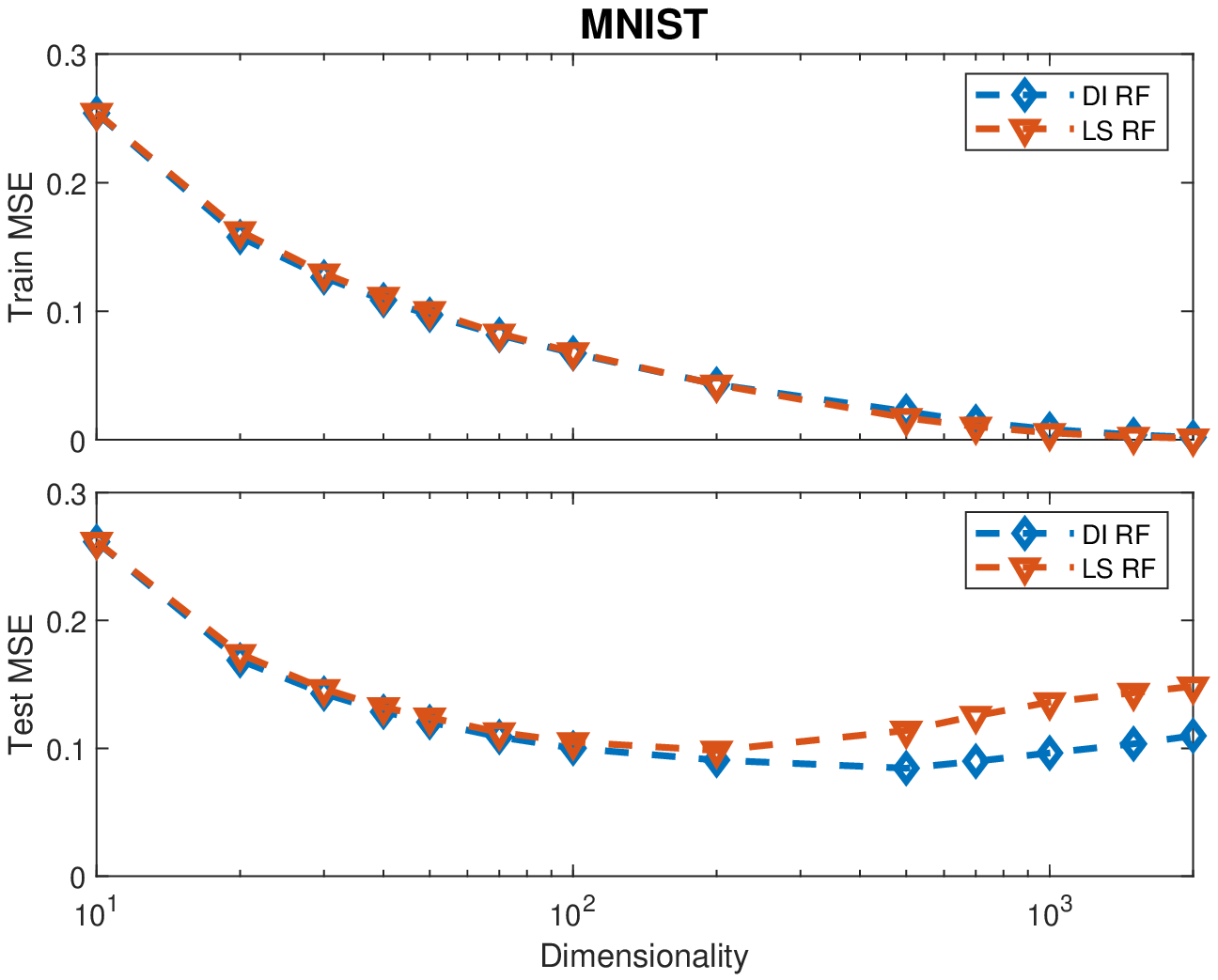}}
\end{minipage}

\hspace{1.3cm}
\begin{minipage}[t]{0.1\textwidth}
  \centering
  \centerline{\includegraphics[width=4.5cm]{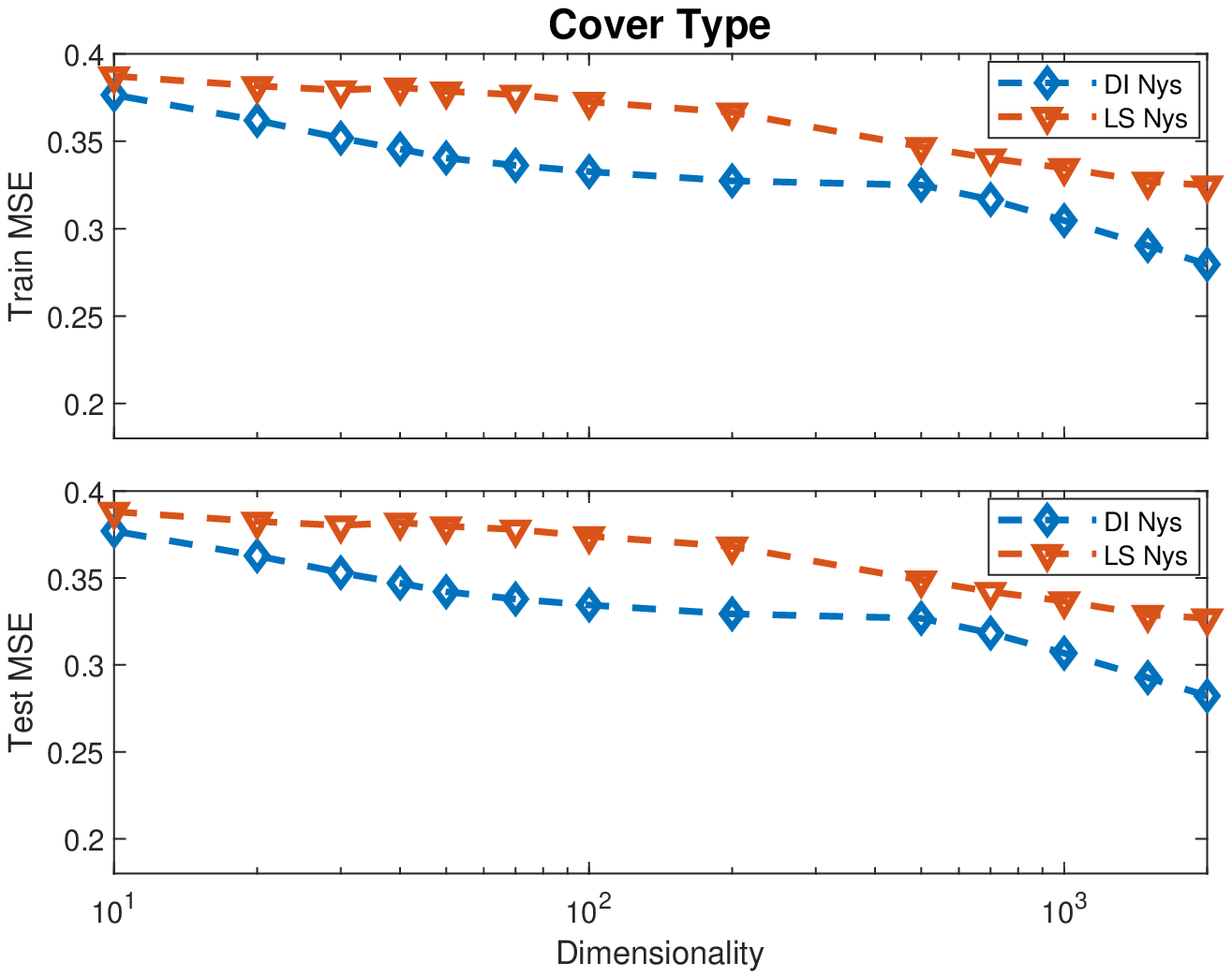}}
\end{minipage}
\hspace{2.5cm}
\begin{minipage}[t]{0.1\textwidth}
  \centering
  \centerline{\includegraphics[width=4.5cm]{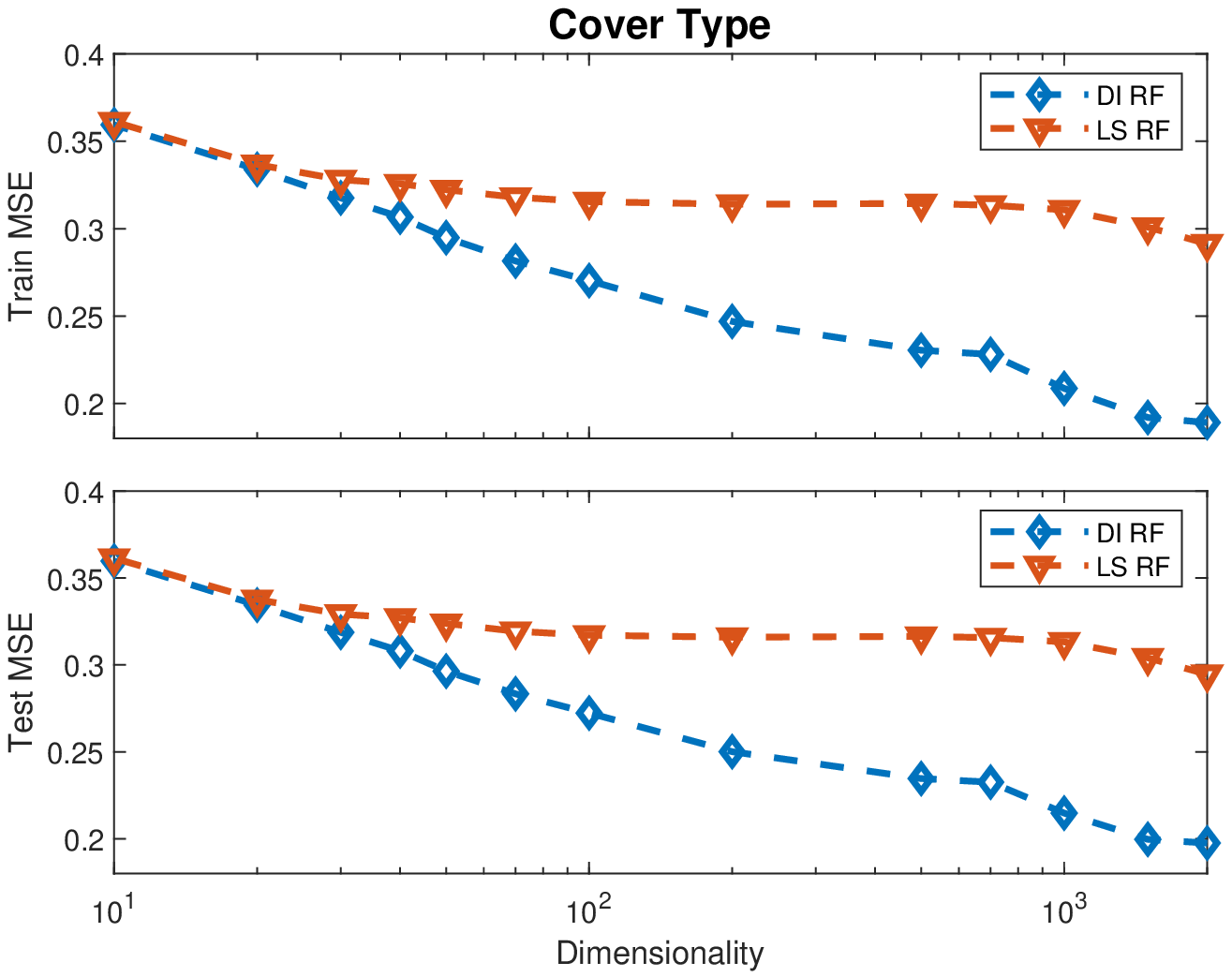}}
\end{minipage}
\vspace{-10pt}
\caption{Training and testing mean square errors (MSEs) achieved by KRR after DI and LS based kernel optimization. }
\label{fig:MSEs}
\vspace{-5pt}
\end{figure}

The datasets used in our experiments are summarized in Table \ref{tab:Datasets}. 
We scaled the feature values to be in $[0,1]$ and used Gaussian (RBF) kernels, i.e.,
$K(\bx_i,\bx_j) = \exp( -\gamma \left\| \bx_i-\bx_j \right\|_2^2 )$, for all the experiments. The kernel parameters ($\gamma$) were pre-determined via 3-fold cross-validation grid search utilizing Kernel SVM with the standard Nystr\"om approximation using $1000$-dimensional feature maps. 

We found that the performances of the trained kernels are robust to changes in the value of the ridge regularizer $\rho$ as long as this parameter is sufficiently small. Therefore, we set $\rho=10^{-4}$ throughout our experiments. We also found the choice between random sampling and k-means initializations to have minimal impact on the quality of the optimized Nystr\"om kernels. For this reason, we simply initialize $\bX_r$ to be a random subset of training samples. For Random Fourier features, our kernel choice leads to initial $\bW_f$ being sampled from the Gaussian distribution with mean $0$ and variance $2\gamma$. 

We achieved the most competitive optimization performances for the evaluated methods by utilizing the Adam optimizer \cite{kingma2014adam} with stochastic gradients and decaying learning rates. Accordingly, we set the initial learning rate to $10^{-3}$ and multiply it with a factor of $10^{-1}$ as the objective value saturates. We stop the training when the loss fails to decline after a learning rate decay. We set the batch size to 1000 throughout our experiments. For DI based training with feature dimensionalities greater than 500, we distinctly set the batch size to twice the feature dimensionality as a regularizer. 

We perform two evaluations on the DI based kernel learning methodology. We first compare the training and generalization performances of our method to the Least Squares (LS) based kernel learning methodology as presented in \cite{si2016goal}. Next, we evaluate DI's effectiveness in learning good kernels for classification tasks. The reported values are produced by averaging over 10 random experiments.
\vspace{-5pt}
\subsection{Comparison of DI and LS Based Training}
\vspace{-2pt}
\begin{figure}[t]
\hspace{1.3cm}
\begin{minipage}[t]{0.1\textwidth}
  \centering
  \centerline{\includegraphics[width=4.5cm]{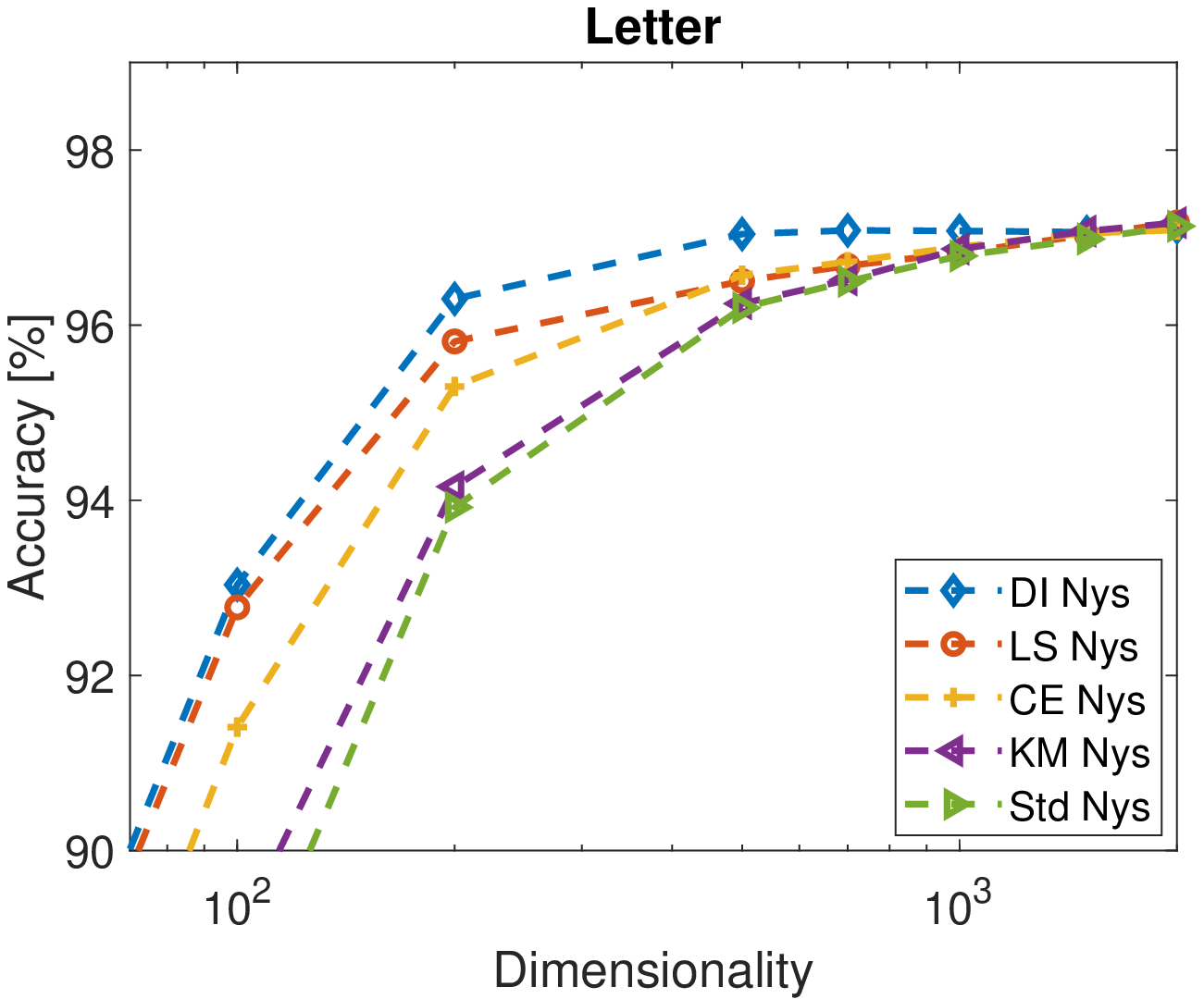}}
\end{minipage}
\hspace{2.5cm}
\begin{minipage}[t]{0.1\textwidth}
  \centering
  \centerline{\includegraphics[width=4.5cm]{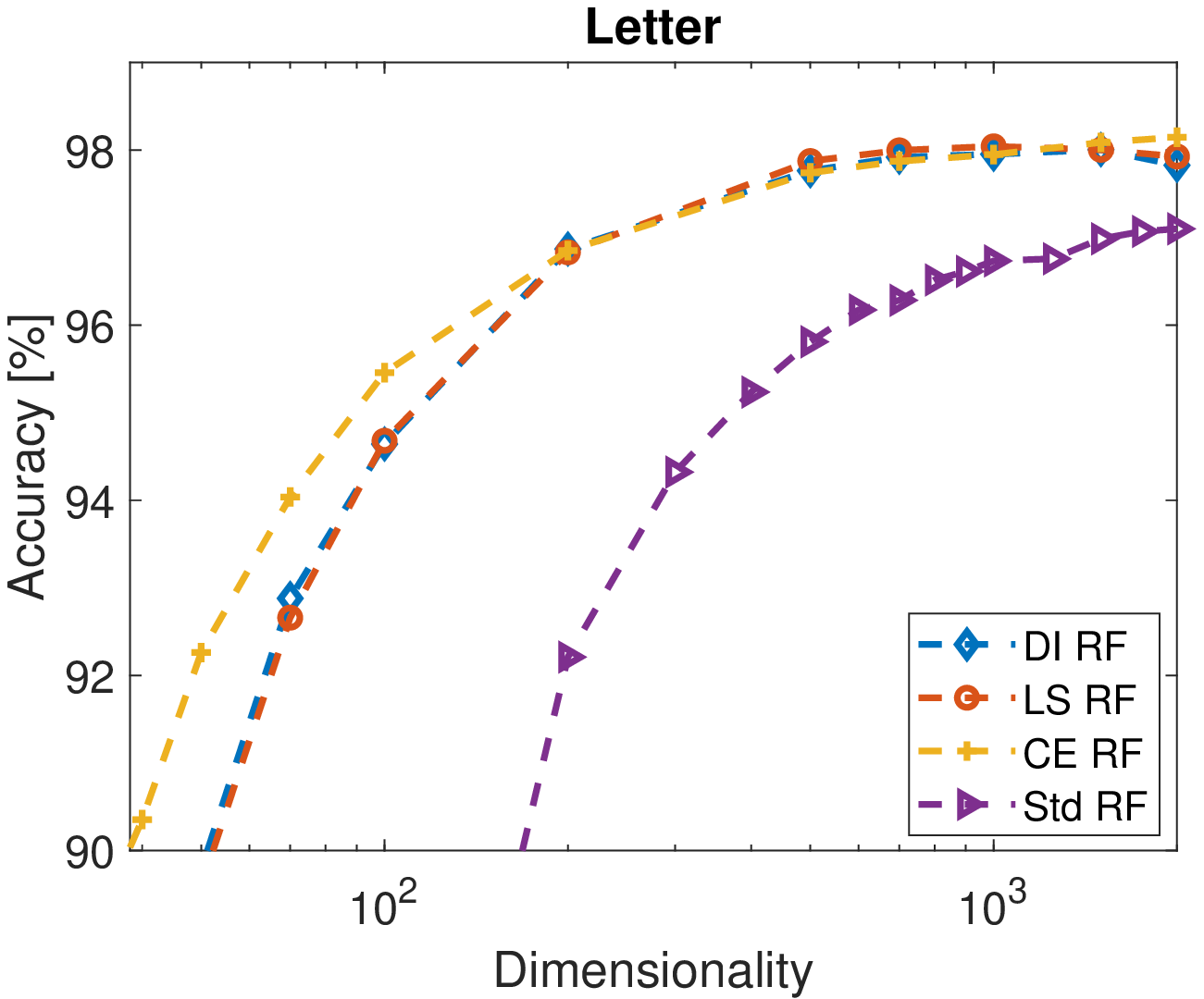}}
\end{minipage}

\hspace{1.3cm}
\begin{minipage}[t]{0.1\textwidth}
  \centering
  \centerline{\includegraphics[width=4.5cm]{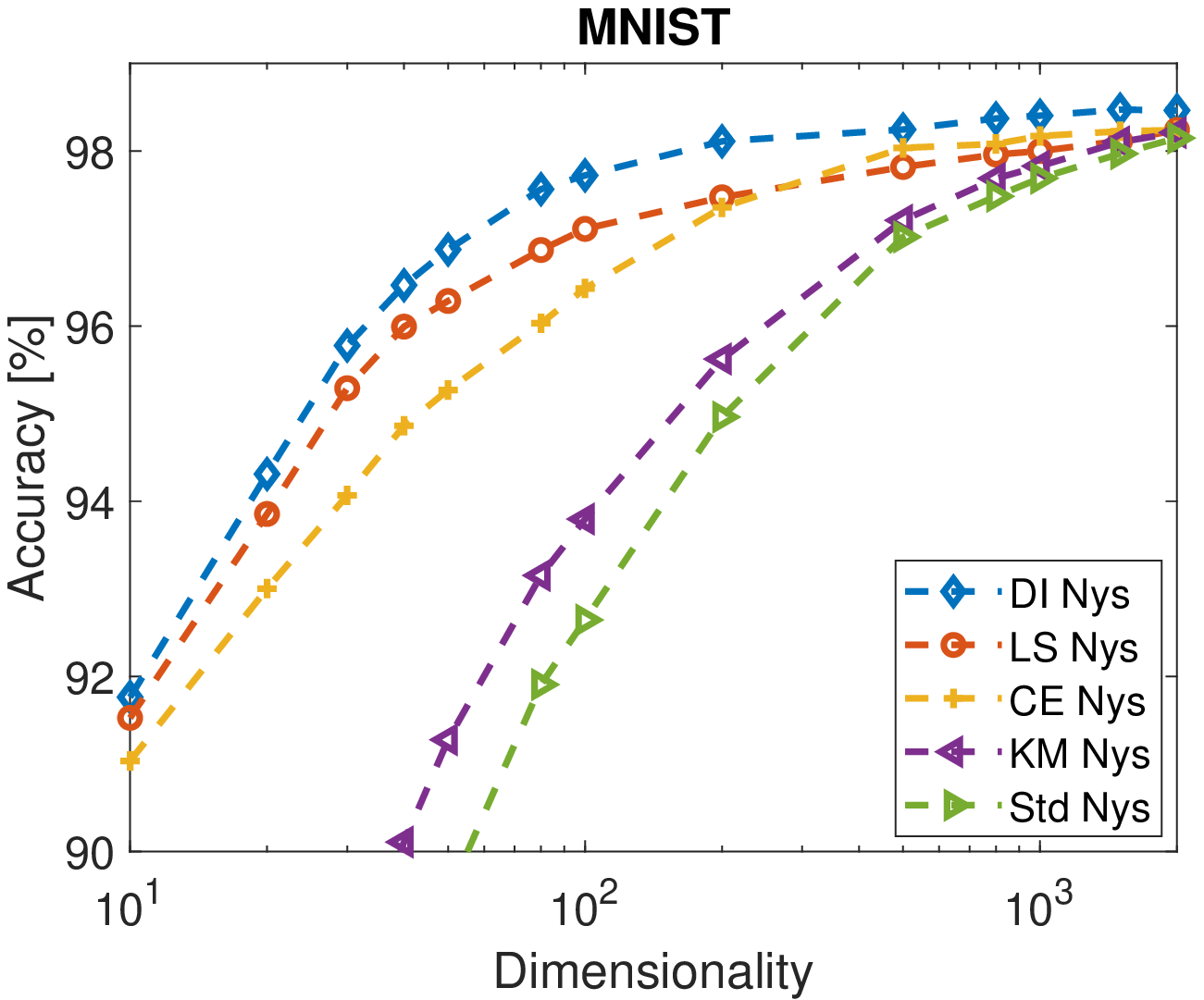}}
\end{minipage}
\hspace{2.5cm}
\begin{minipage}[t]{0.1\textwidth}
  \centering
  \centerline{\includegraphics[width=4.5cm]{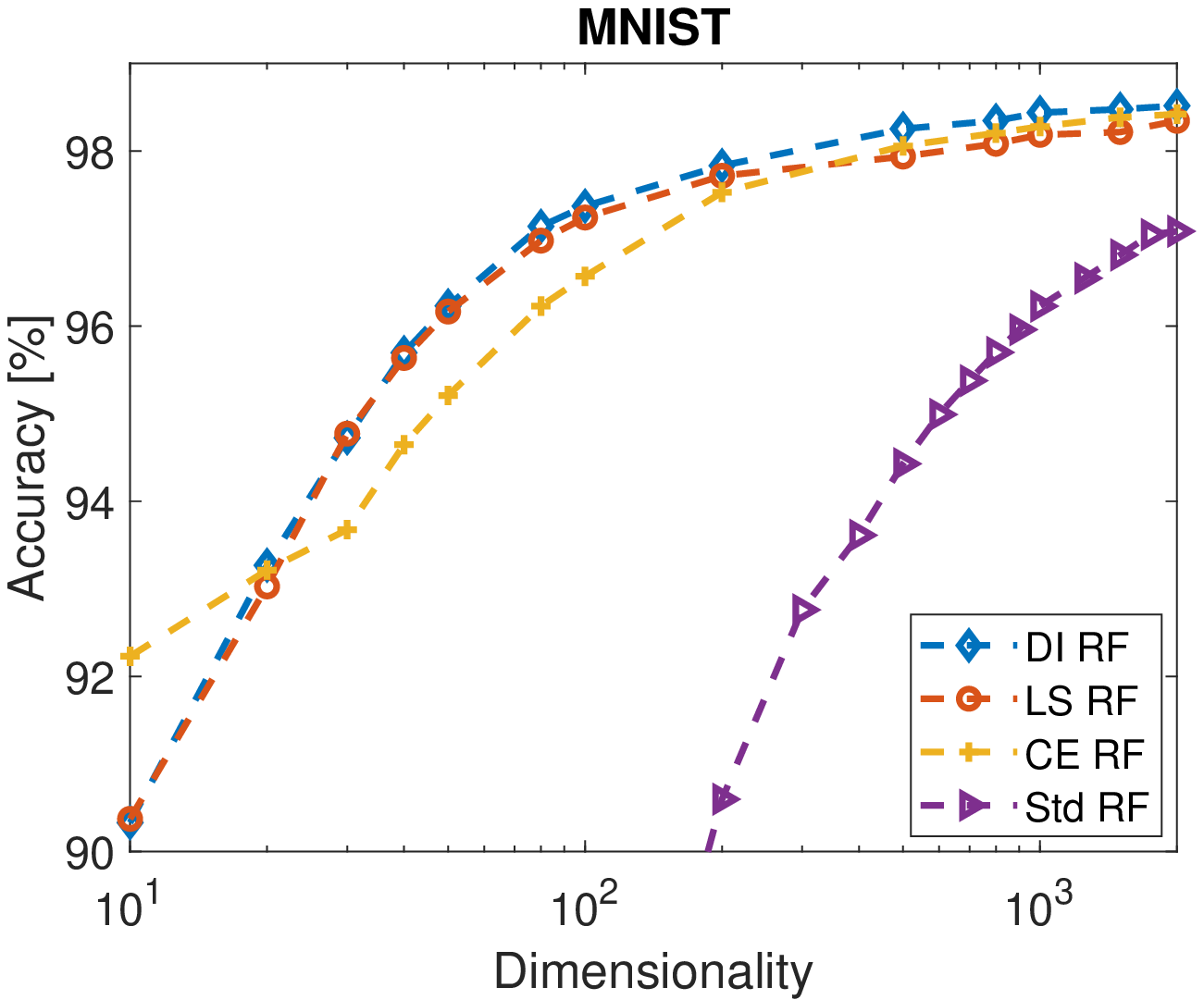}}
\end{minipage}

\hspace{1.3cm}
\begin{minipage}[t]{0.1\textwidth}
  \centering
  \centerline{\includegraphics[width=4.5cm]{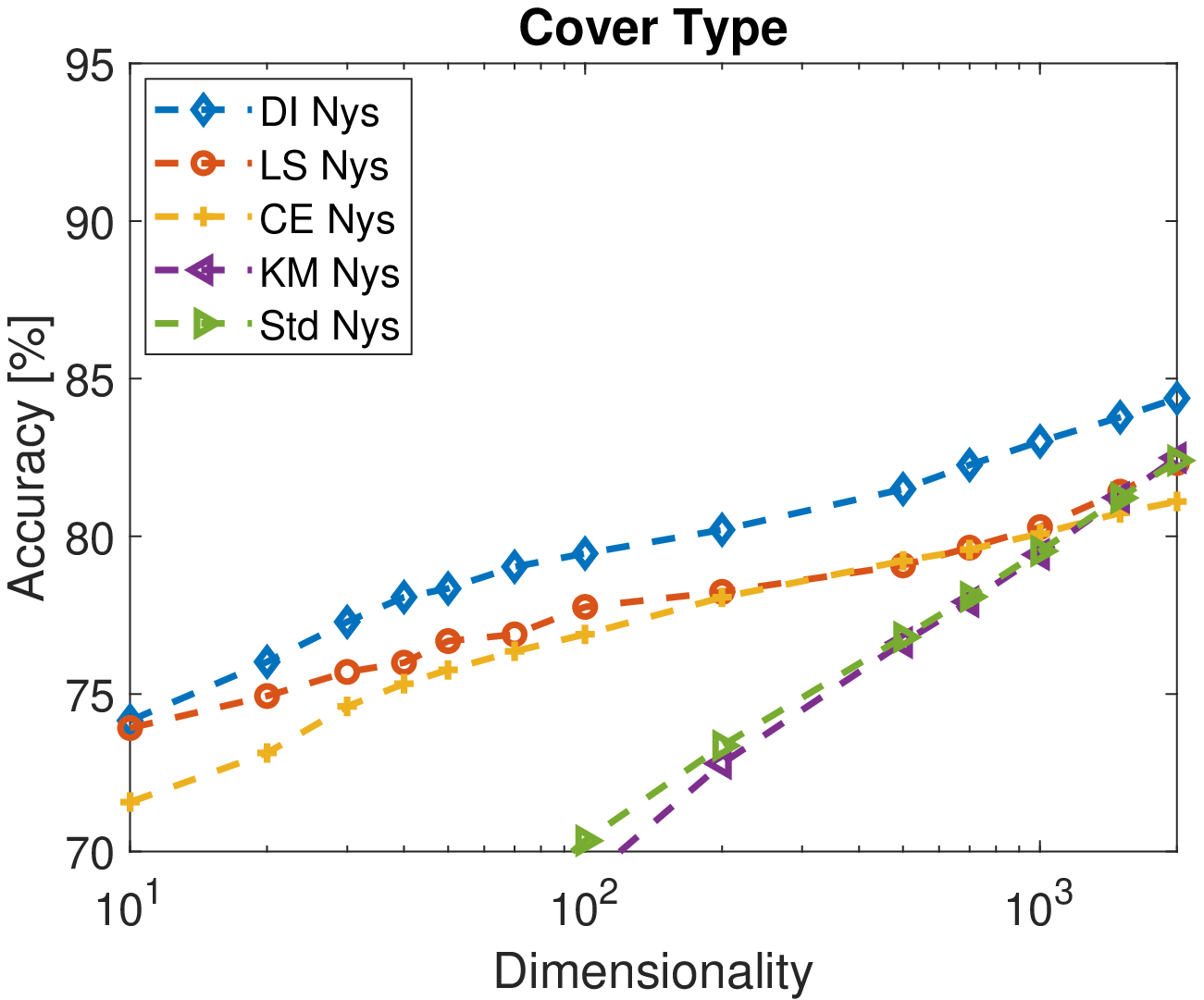}}
\end{minipage}
\hspace{2.5cm}
\begin{minipage}[b]{0.1\textwidth}
  \centering
  \centerline{\includegraphics[width=4.5cm]{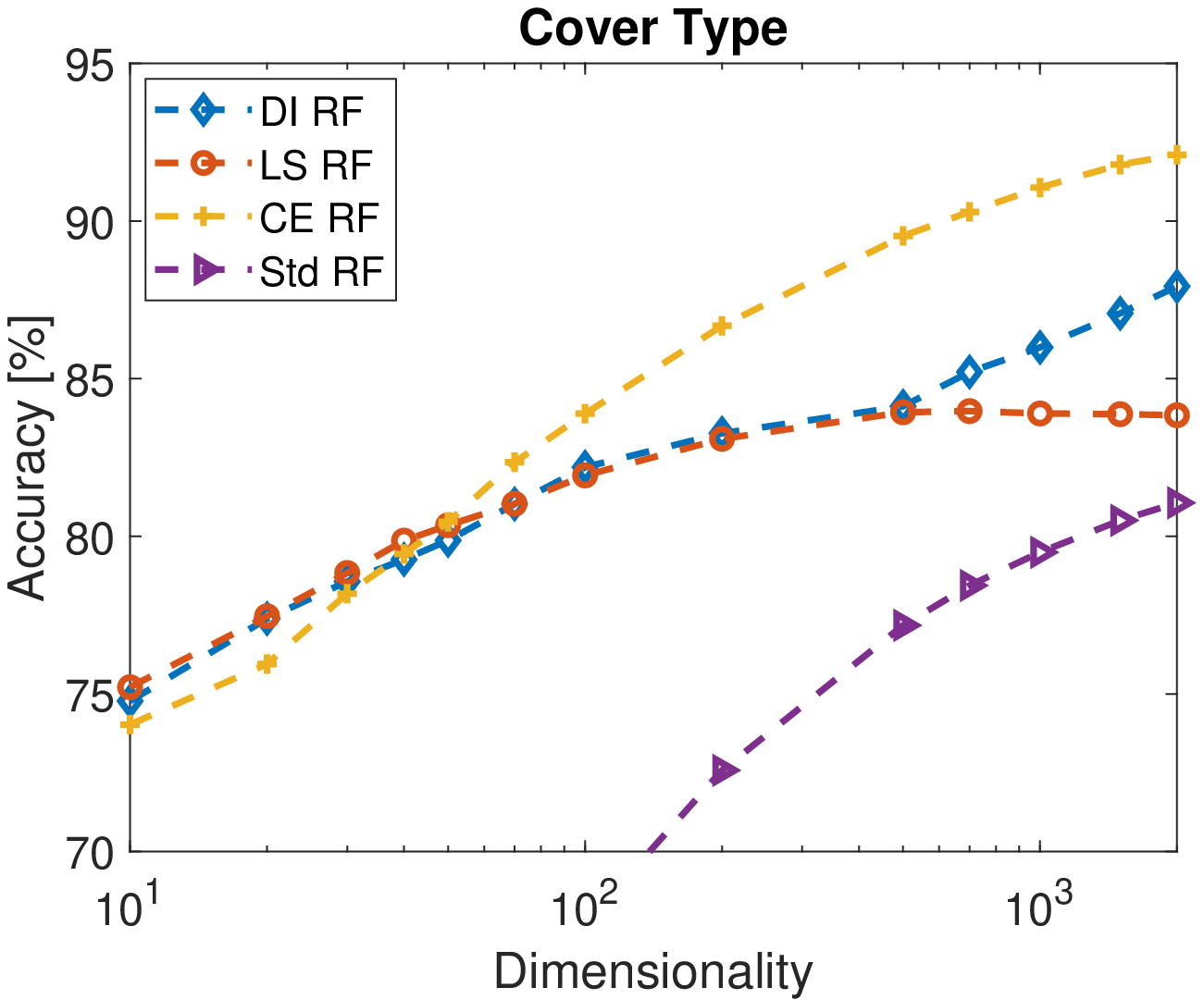}}
\end{minipage}
\vspace{-10pt}
\caption{The prediction accuracies of SVMs trained after DI, LS and CE based kernel optimization as well as the standard kernel approximation techniques. 
}
\label{fig:Accuracies}
\vspace{-5pt}
\end{figure}

In Section \ref{subsec:DI}, we established the connection between DI and LS Regression. Indeed, full-batch DI based training would search for a feature map that best minimizes the Ridge Regression (RR) loss on the training set. However, since each mini-batch represents a different empirical distribution, mini-batch DI based training can yield feature maps that adapt to variations in the empirical distribution.

We compare our method to an algorithm presented in \cite{si2016goal}, where an RR predictor is trained jointly with the kernel. The implementation alternates between optimizing the RR weights and taking a gradient step over the parameters of the kernel mapping w.r.t. the RR loss. Since this methodology explicitly optimizes Nys and RF features to minimize the RR loss, it forms the LS baseline, and it is of especial interest to compare it to our DI based kernel learning approach.

Figure \ref{fig:MSEs} displays the train and test mean squared errors (MSEs) achieved by KRR, when LS and DI based training is performed on the kernel. With Nystr\"om features, DI based training consistently leads to better optimization performances on the training sets, an observation we also make on CovType with Random Fourier features. This is despite the fact that LS based training explicitly minimizes MSE on the training set and both methods are trained until saturation.

These results indicate that DI based training with stochastic gradients is capable of leading to better solutions. This seems especially true while training representative data points for Nystr\"om features, which, due to our kernel choice, may otherwise suffer from structural difficulties associated with training RBF layers \cite{lecun1998gradient}.

DI and LS based training achieve very similar MSEs with Random Fourier features on Letter and MNIST. However, DI based training significantly outperforms on the test set of MNIST at high feature dimensionalities. While both DI and LS trained Random Fourier features start overfitting the training set with increased dimensionalities, we observe overfitting later and at a lesser extent with DI training. This improved generalization performance may be explained by DI training effectively being performed over a large ensemble of linear predictors optimized over separate mini-batches.
\vspace{-5pt}
\subsection{Evaluation of DI for Classification}
\vspace{-2pt}
In this section, we evaluate how DI optimized kernels perform in classification tasks. To this end, we augment the LS baseline in the previous section with a margin maximizing training objective, namely, the softmax cross-entropy (CE) loss. 
We also report the accuracies achieved by standard Nystr\"om (Std Nys), K-Means Nystr\"om (KM Nys) and standard Random Fourier features (Std RF).

We evaluate the kernels by training SVMs on the learned feature maps and reporting the prediction accuracies, which are presented in Figure \ref{fig:Accuracies}. We see that DI based optimization yields significantly better results with Nystr\"om features and generally outperforms LS based optimization with both types of features. However, it under-performs compared to CE training with Random Fourier features on Letter and CovType data. Considering the structure and initialization of Random Fourier features resemble a hidden layer, this result is in line with CE being a popular objective function for Neural Network training.

Overall, we observe that DI tends to be a better alternative to LS for training kernel mappings and can also outperform CE in some settings, as demonstrated by our results from MNIST and the Nystr\"om experiments. Hence, DI based optimization proves to be suitable for learning kernels with high discriminant power, with CE remainining as a good alternative for training kernel maps whose structures resemble a Neural Network.

\vspace{-5pt}
\section{Discussion}
\vspace{-2pt}
This paper considers DI for optimizing two particular types of kernel mappings. Nevertheless, the DI based training procedure can in principle be applied to any parametric non-linear mapping. This includes such kernel approximation methods as \emph{Fastfood Expansions} \cite{le2013fastfood,yang2015carte}, which enable faster processing of datasets with high input dimensions.
The kernel mappings can also be combined with other elements, e.g., convolutional or recursive networks, which would be better suited for application areas such as image and speech processing. 

Additionally, DI/KDI maximizing kernels can be applied for purposes other than supervised learning. For example, Eric \textit{et al.} utilize the Kernel Fisher Discriminant Ratio, a specific instance of the proposed Kernel DI, in two sample tests of homogeneity as an alternative to the commonly used measure called Maximum Mean Discrepancy (MMD) \cite{eric2008testing}. Thus, maximal DI over a class of kernels can potentially serve as a generative model training objective following the usage of maximal MMD for training generative networks \cite{li2017mmd}.

\vspace{-5pt}
\section{Conclusion}
\vspace{-2pt}
We have proposed a novel methodology for training low-dimensional kernel mappings to improve their performances at supervised learning tasks. The proposed Discriminant Information objective is suitable for a wide range of kernel maps and can successfully improve the optimization and generalization performances over existing kernel optimization techniques. In the future, we hope to extend our kernel learning methodology to more general objectives that may not allow for closed form expressions of the minimum loss. In addition, we plan to utilize our methodology with an extended class of feature mappings and learning settings for added versatility.

\bibliographystyle{IEEEbib}
\bibliography{refs}

\begin{thebibliography}{10}

\bibitem{vapnik2013nature}
Vladimir Vapnik,
\newblock {\em The nature of statistical learning theory},
\newblock Springer, Verlag, NY, USA, 2nd edition, 2000.

\bibitem{kung2014kernel}
Sun-Yuan Kung,
\newblock {\em Kernel methods and machine learning},
\newblock Cambridge University Press, New York, NY, USA, 2014.

\bibitem{girolami2002orthogonal}
Mark Girolami,
\newblock ``Orthogonal series density estimation and the kernel eigenvalue
  problem,''
\newblock {\em Neural Computation}, vol. 14, no. 3, pp. 669--688, Mar. 2002.

\bibitem{drineas2005nystrom}
Petros Drineas and Michael~W Mahoney,
\newblock ``On the {N}ystr{\"o}m method for approximating a gram matrix for
  improved kernel-based learning,''
\newblock {\em Journal of Machine Learning Research}, vol. 6, pp. 2153--2175,
  Dec. 2005.

\bibitem{rahimi2008random}
Ali Rahimi and Benjamin Recht,
\newblock ``Random features for large-scale kernel machines,''
\newblock in {\em Proc.~Advances in Neural Information Processing Systems},
  Dec. 2008, pp. 1177--1184.

\bibitem{rahimi2009weighted}
Ali Rahimi and Benjamin Recht,
\newblock ``Weighted sums of random kitchen sinks: Replacing minimization with
  randomization in learning,''
\newblock in {\em Proc.~Advances in Neural Information Processing Systems},
  Dec. 2009, pp. 1313--1320.

\bibitem{yang2015carte}
Zichao Yang, Andrew Wilson, Alex Smola, and Le~Song,
\newblock ``A la carte--learning fast kernels,''
\newblock in {\em Proc.~Artificial Intelligence and Statistics}, May 2015, pp.
  1098--1106.

\bibitem{si2016goal}
Si~Si, Kai-Yang Chiang, Cho-Jui Hsieh, Nikhil Rao, and Inderjit~S Dhillon,
\newblock ``Goal-directed inductive matrix completion,''
\newblock in {\em Pro.~ACM SIGKDD International Conference on Knowledge
  Discovery and Data Mining}. ACM, Aug. 2016, pp. 1165--1174.

\bibitem{dai2014scalable}
Bo~Dai, Bo~Xie, Niao He, Yingyu Liang, Anant Raj, Maria-Florina~F Balcan, and
  Le~Song,
\newblock ``Scalable kernel methods via doubly stochastic gradients,''
\newblock in {\em Advances in Neural Information Processing Systems}, 2014, pp.
  3041--3049.

\bibitem{lu2014scale}
Zhiyun Lu, Avner May, Kuan Liu, Alireza~Bagheri Garakani, Dong Guo,
  Aur{\'e}lien Bellet, Linxi Fan, Michael Collins, Brian Kingsbury, Michael
  Picheny, et~al.,
\newblock ``How to scale up kernel methods to be as good as deep neural nets,''
\newblock {\em arXiv preprint arXiv:1411.4000}, 2014.

\bibitem{kung2017compressive}
Sun-Yuan Kung,
\newblock ``Compressive privacy: From information/estimation theory to machine
  learning [lecture notes],''
\newblock {\em IEEE Signal Processing Magazine}, vol. 34, no. 1, pp. 94--112,
  2017.

\bibitem{zhang2008improved}
Kai Zhang, Ivor~W Tsang, and James~T Kwok,
\newblock ``Improved {N}ystr{\"o}m low-rank approximation and error analysis,''
\newblock in {\em Proc.~International Conference on Machine Learning}. ACM,
  Jul. 2008, pp. 1232--1239.

\bibitem{kumar2009ensemble}
Sanjiv Kumar, Mehryar Mohri, and Ameet Talwalkar,
\newblock ``Ensemble {N}ystr{\"o}m method,''
\newblock in {\em Proc.~Advances in Neural Information Processing Systems},
  Dec. 2009, pp. 1060--1068.

\bibitem{li2015large}
Mu~Li, Wei Bi, James~T Kwok, and Bao-Liang Lu,
\newblock ``Large-scale {N}ystr{\"o}m kernel matrix approximation using
  randomized {SVD},''
\newblock {\em IEEE Transactions on Neural Networks and Learning Systems}, vol.
  26, no. 1, pp. 152--164, Jan. 2015.

\bibitem{gittens2016revisiting}
Alex Gittens and Michael~W Mahoney,
\newblock ``Revisiting the {N}ystr{\"o}m method for improved large-scale
  machine learning,''
\newblock {\em Journal of Machine Learning Research}, vol. 17, no. 1, pp.
  3977--4041, Apr. 2016.

\bibitem{le2013fastfood}
Quoc Le, Tam{\'a}s Sarl{\'o}s, and Alex Smola,
\newblock ``Fastfood-approximating kernel expansions in loglinear time,''
\newblock in {\em Proc.~International Conference on Machine Learning}, Jun.
  2013, vol.~85.

\bibitem{mairal2014convolutional}
Julien Mairal, Piotr Koniusz, Zaid Harchaoui, and Cordelia Schmid,
\newblock ``Convolutional kernel networks,''
\newblock in {\em Advances in neural information processing systems}, Dec 2014,
  pp. 2627--2635.

\bibitem{bryan1951generalized}
Joseph~G Bryan,
\newblock ``The generalized discriminant function: mathematical foundation and
  computational routine,''
\newblock {\em Harvard educational review}, vol. 21, no. 2, pp. 90--95, 1951.

\bibitem{roth2000nonlinear}
Volker Roth and Volker Steinhage,
\newblock ``Nonlinear discriminant analysis using kernel functions,''
\newblock in {\em Advances in neural information processing systems}, 2000, pp.
  568--574.

\bibitem{kung2017discriminant}
Sun-Yuan Kung,
\newblock ``Discriminant component analysis for privacy protection and
  visualization of big data,''
\newblock {\em Multimedia Tools and Applications}, vol. 76, no. 3, pp.
  3999--4034, 2017.

\bibitem{friedman2001elements}
Jerome Friedman, Trevor Hastie, and Robert Tibshirani,
\newblock {\em The elements of statistical learning}, vol.~1,
\newblock Springer, Verlag, NY, USA, 2nd edition, 2001.

\bibitem{mao2002rbf}
K.~Z. {Mao},
\newblock ``Rbf neural network center selection based on fisher ratio class
  separability measure,''
\newblock {\em IEEE Transactions on Neural Networks}, vol. 13, no. 5, pp.
  1211--1217, Sep. 2002.

\bibitem{eric2008testing}
Moulines Eric, Francis~R Bach, and Za{\"\i}d Harchaoui,
\newblock ``Testing for homogeneity with kernel fisher discriminant analysis,''
\newblock in {\em Advances in Neural Information Processing Systems}, Dec.
  2008, pp. 609--616.

\bibitem{frey1991letter}
Peter~W Frey and David~J Slate,
\newblock ``Letter recognition using holland-style adaptive classifiers,''
\newblock {\em Machine Learning}, vol. 6, no. 2, pp. 161--182, Mar. 1991.

\bibitem{lecun1998gradient}
Yann LeCun, L{\'e}on Bottou, Yoshua Bengio, and Patrick Haffner,
\newblock ``Gradient-based learning applied to document recognition,''
\newblock {\em Proceedings of the IEEE}, vol. 86, no. 11, pp. 2278--2324, Nov.
  1998.

\bibitem{blackard1999comparative}
J~A Blackard and D~J Dean,
\newblock ``Comparative accuracies of artificial neural networks and
  discriminant analysis in predicting forest cover types from cartographic
  variables,''
\newblock {\em Computers and Electronics in Agriculture}, vol. 24, pp.
  131--151, 1999.

\bibitem{kingma2014adam}
Diederik~P Kingma and Jimmy Ba,
\newblock ``Adam: A method for stochastic optimization,''
\newblock {\em arXiv preprint arXiv:1412.6980}, 2014.

\bibitem{li2017mmd}
Chun-Liang Li, Wei-Cheng Chang, Yu~Cheng, Yiming Yang, and Barnab{\'a}s
  P{\'o}czos,
\newblock ``Mmd gan: Towards deeper understanding of moment matching network,''
\newblock in {\em Advances in Neural Information Processing Systems}, 2017, pp.
  2203--2213.

\end{thebibliography}

\end{document}